\documentclass[letterpaper]{article}
\usepackage{iccc}
\usepackage{times}
\usepackage{helvet}
\usepackage{courier}
\usepackage{url}            
\usepackage{booktabs}       
\usepackage{amsfonts}       
\usepackage{nicefrac}       
\usepackage{microtype}      
\usepackage{subcaption}
\usepackage{multicol}
\usepackage{amsmath}
\usepackage{graphicx}
\usepackage{stfloats}

\title{Beyond Imitation: \\ Generative and Variational Choreography via Machine Learning\\}

\author{
  Mariel Pettee \\
  Department of Physics\\ 
  Yale University \\
  New Haven, CT 06511\\
  mariel.pettee@yale.edu \\
  \And
  Chase Shimmin \\
  Department of Physics\\ 
  Yale University \\
  New Haven, CT 06511\\
  chase.shimmin@yale.edu \\
  \And
  Douglas Duhaime \\
  Digital Humanities Laboratory\\
  Yale University \\
  New Haven, CT 06511\\
  douglas.duhaime@yale.edu \\
  \And
  Ilya Vidrin \\
  Coventry University \\Centre for Dance Research\\
  Harvard University Dance Program\\
  Cambridge, MA 02138\\
  ilya\_vidrin@mail.harvard.edu \\
}
\begin{document} 
\maketitle

\begin{abstract}
\begin{quote}
Our team of dance artists, physicists, and machine learning researchers has collectively developed several original, configurable machine-learning tools to generate novel sequences of choreography as well as tunable variations on input choreographic sequences. We use recurrent neural network and autoencoder architectures from a training dataset of movements captured as 53 three-dimensional points at each timestep. Sample animations of generated sequences and an interactive version of our model can be found at \url{http://www.beyondimitation.com}.
\end{quote}
\end{abstract}

\section{Introduction}
``I didn't want to imitate anybody. Any movement I knew, I didn't want to use.'' \cite{pina} Eminent postmodern dance choreographer Pina Bausch felt the same ache that has pierced artists of all generations -- the desire to generate something truly original from within the constraints of your own body.

Recent technologies enabling the 3D capture of human motion as well as the analysis and prediction of timeseries datasets with machine learning have opened provocative new possibilities in the domain of movement generation. In this paper, we introduce a suite of configurable machine learning tools to augment a choreographer's workflow.

Many generative movement models from recent publications use Recurrent Neural Networks (RNNs) \cite{graves} as their fundamental architecture \cite{mccormick,chorrnn,alemi,zhou,berman,markovic}. Others create methods to draw trajectories through a lower-dimensional space of possible human poses constructed through techniques such as Kernel Principal Component Analysis (KPCA) \cite{james,kpca}. We build upon existing RNN techniques with higher-dimensional datasets and introduce autoencoders \cite{ae} of both poses and sequences of poses to construct variations on input sequences of movement data and novel unprompted sequences sampled from a lower-dimensional latent space.

 Our models not only generate new movements and dance sequences both with and without a movement prompt, but can also create infinitely many variations on a given input phrase. These methods have been developed using a dataset of improvisational dance from one of the authors herself, recorded using a state-of-the-art motion capture system with a rich density of datapoints representing the human form. With this toolset, we equip artists and movement creators with strategies to tackle the challenge Bausch faced in her own work: generating truly novel movements with both structure and aesthetic meaning.

\subsection{Context within Dance Scholarship} 

Dance scholarship, psychology, and philosophy of the past century has increasingly seen movement as embodied thought. Prominent proposals including psychologist Jean Piaget's sensorimotor stage of psychological development, the philosopher Maurice Merleau-Ponty's ``phenomenology of embodiment'', and Edward Warburton's concept of \emph{dance enaction} have guided us today to view the human body as an essential influencer of cognition and perception \cite{warburton}. 

Our team's vision for the future of creative artificial intelligence necessitates the modeling of not only written, visual, and musical thought, but also kinesthetic comprehension. In light of the modern understanding of movement as an intellectual discipline, the application of machine learning to movement research serves not as a mere outsourcing of physical creative expressiveness to machines, but rather as a tool to spark introspection and exploration of our embodied knowledge.

\begin{figure}[h]
\centering
\begin{minipage}{\linewidth}
  \includegraphics[width=\linewidth]{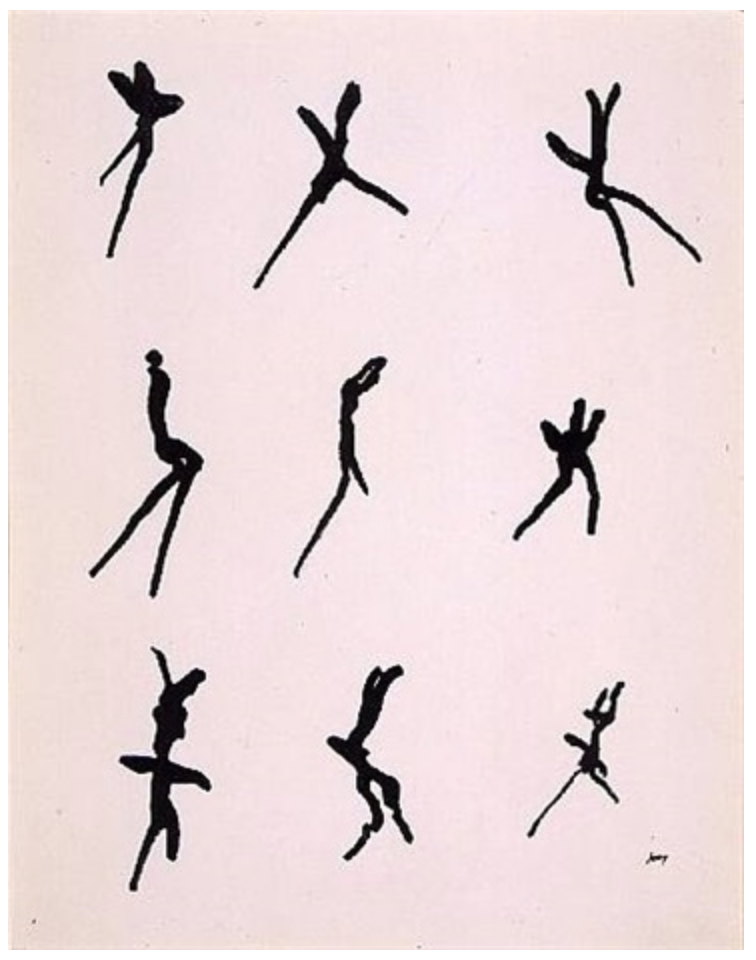}
  \captionof{figure}{ Henri Michaux's notion of \emph{envie cin\'etique}, or ``kinetic desire'', is represented by expressive, personal, and idiosyncratic gestures in calligraphic ink in his series \emph{Mouvements} \cite{noland}.}
  \label{fig:mouvements}
\end{minipage}
\end{figure}

Concurrently with this branch of research, choreographers have wrestled with the problem of constructing a universal language of movement. Movement writing systems currently in use today such as Labanotation, Benesh Choreology, and Eshkol-Wachmann Notation can be effective in limited use cases, but make culturally-specific assumptions about how human bodies and types of motion should be abstracted and codified \cite{farnell}. 

It is not our aim to replace existing methods of dance notation. However, we note the significance of 3D motion-capture techniques and abstract latent spaces in potentially reorienting movement notation away from culturally-centered opinions such as qualities of movement or which segments of the body get to define movement. Rather than gravitating in the direction of defining ``universal'' movement signifiers, we see this work as more aligned with the expressive figures generated by the visual artist Henri Michaux in an attempt to capture what he called \emph{envie cin\'etique}, or ``kinetic desire'' -- in other words, the pure impulse to move (see Figure \ref{fig:mouvements}). We therefore avoid limiting our generated movement outputs to only physically-achievable gestures, as this would only serve to limit the potential imaginative sparks lying dormant in these sequences. 

Ethics in the philosophy of emerging media raise particular questions about how technology impacts what it means to be human, especially given the way constraints and resources of technology affect our embodied dispositions. When we consider the ethical dimensions of choreography in the context of machine learning, one major benefit is the opportunity to reflect on habits by observing, interpreting, and evaluating what is generated technologically. The normative problems that ensue are manifold: if we ascribe great value to what we see, we may find ourselves in a position where we envy an algorithm's capacity to generate novel choreography. This may in turn lead us to cast judgement on ourselves and doubt our own human-created choreographies. 

While technology may provide new insights into patterns within dance sequences, it also inevitably leads to normative discussion about what it means to choreograph well, or appropriately, or even creatively. This opens the door for replacing our own practice with algorithms that could ostensibly rob us of the opportunity to get better at choreography, or learn to be more creative. While this may seem a bit like catastrophizing, these normative problems can lead to real ethical concerns related not only to artistic practice, but to education more broadly. 

Several prominent choreographers have sought out both motion capture and machine learning tools to augment their practice, from Bill T. Jones and the OpenEndedGroup's 1999 motion capture piece \emph{Ghostcatching} to William Forsythe to Merce Cunningham \cite{ghostcatching,forsythe,cunningham}. Wayne McGregor recently collaborated with Google Arts \& Culture to create \emph{Living Archive}, a machine learning-based platform to generate a set of movements given an input sequence derived from a video, although details of the technical implementation of this project are not yet publicly released \cite{mcgregor}. 

Our work represents a unique direction in the space of ``AI-generated'' choreographies, both computationally and artistically. Computationally, we combine high-dimensional and robust 3D motion capture data with  existing RNN-based architectures as well as introducing the use of autoencoders for 3D pose and movement sequence generation. Artistically, we deviate from having novel predicted sequences as the only end goal -- in addition to this functionality, we grant choreographers the power to finely-tune existing movement sequences to find subtle (or not-so-subtle) variations from their original ideas.

\section{Methods}
Training data was recorded in a studio equipped with 20 Vicon Vantage motion-capture cameras and processed with Vicon Shogun software. This data consists of the positions of 53 fixed vertices on a dancer in 3 dimensions through a series of nearly 60,000 temporal frames recorded at 35 fps, comprising approximately 30 minutes of real-time movement. Each frame of the dataset is transformed such that the overall average (x,y) position per frame is centered at the same point and scaled such that all of the coordinates fit within the unit cube. The primary author, who has an extensive background in contemporary dance, supplied the training data. The data was then exported to Numpy array format for visualization and processing in Python, and to JSON format for visualization with the interactive 3D Javascript library \texttt{three.js}. The neural network models were constructed using Keras with a Tensorflow backend.

In the following subsections, we describe two methods for generating dance movement in both conditional (where a prompt sequence of fixed length is provided) and unconditional (where output is generated without input) modes.
The first method involves a standard approach to supervised training for sequence generation: an RNN is presented with a sequence of training inputs, and is trained to predict the next frame(s) in the sequence.
The second method takes advantage of autoencoders to convert either an arbitrary-length sequence of dance movement into a trajectory of points in a low-dimensional latent space, or a fixed-length sequence to a single point in a higher-dimensional latent space.

\subsection{LSTM+MDN}
The model proposed in \emph{chor-rnn} \cite{chorrnn} uses RNNs to generate dance from a dataset of 25 vertices captured with a single Kinect device, which requires the dancer to remain mostly front-facing in order to capture accurate full-body data. Our RNN model uses an input layer of size $(53\times3\times m)$ to represent 53 three-dimensional vertices with no rotational restrictions in a prompt sequence of $m$ frames at a time. These sequences are then input to a series of LSTM layers, typically three, followed by a Mixture Density Network \cite{alemi} (see Appendix A) which models proposals for the vertex coordinates of the subsequent $n$ frames. The LSTM layers ensure the model is capable of capturing long-term temporal dependencies in the training data, while the MDN layer ensures generated sequences are dynamic and do not stagnate on the conditional average of previous vertex sequences \cite{bishop}. The network is trained using supervised pairs of sequences by minimizing the negative log likelihood (NLL) of the proposed mixture model. 

We also developed a modification of this structure using Principal Component Analysis (PCA) to reduce the dimensionality of the input sequences. This reduces the amount of information that must be represented by each LSTM layer. We then invert the PCA transformation to convert generated sequences in the reduced-dimensional space back into the $(53\times 3\times n)$-dimensional space.

\begin{figure}[h]
\centering
\begin{minipage}{\linewidth}
  \includegraphics[width=\linewidth]{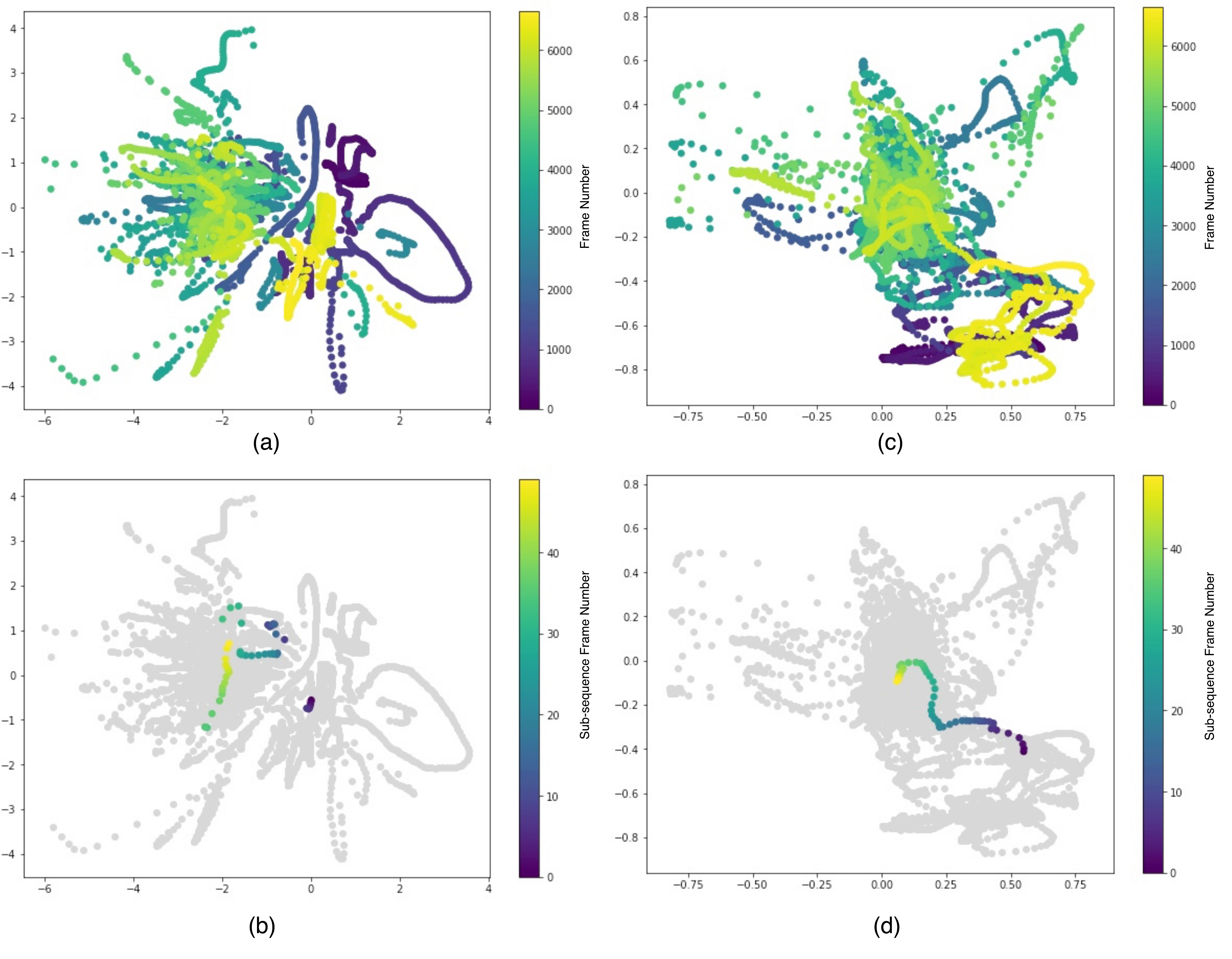}
  \captionof{figure}{\textbf{(a)} The 2-dimensional latent space of an autoencoder trained on a subset of the full dataset. The frame numbers show the procession of the sequence through time at a frame rate of 35 fps. \textbf{(b)} An example sequence of real training data is highlighted in this latent space. Note that its structure is highly noncontinuous. \textbf{(c)} The 2-dimensional latent space of an autoencoder trained on the same subset of data as the previous plots, but with the angular orientation of the frames subtracted. \textbf{(d)} The same sequence of real training data is highlighted, showing a much smoother and more continuous structure.}
  \label{fig:ae_comparison}
\end{minipage}
\end{figure}

\subsection{Autoencoder Methods}
Unlike the RNN methods described above, autoencoders can learn features of the training data with a less directly supervised approach. The input and output layers are identical in dimensionality, while the intermediate layer or layers are of a reduced dimension, creating a characteristic bottleneck shape in the network architecture. The full network is then trained to replicate the training samples as much as possible by minimizing the mean-squared error loss between the input and the generated output. The network therefore learns a reduced dimensionality representation of ``interesting'' features in an unsupervised manner, which can be exploited in the synthesis of new types of movement.

While a well-trained autoencoder merely mimics any input data fed into it, the resulting network produces two useful artifacts: an \emph{encoder} that maps inputs of dimension $(53 \times 3\times m)$ to a $(d\times m)$-dimensional space ($d<159$) and a \emph{decoder} that maps $(d\times m)$-dimensional data back into the original dimensionality of $(53\times 3 \times m)$. This allows us to generate new poses and sequences of poses by tracing paths throughout the $(d\times m)$-dimensional latent space which differ from those found in the training data. 

\begin{figure*}[t!]
\includegraphics[width=0.8\paperwidth]{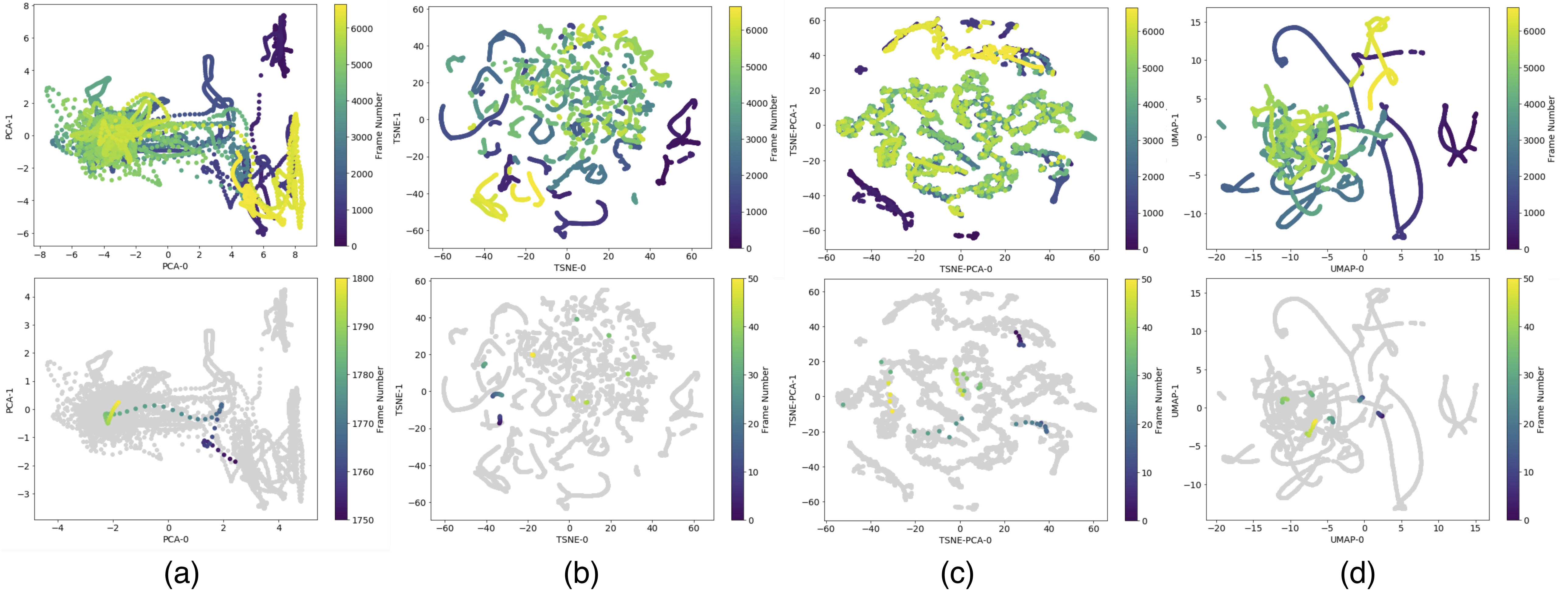}
\caption{A variety of 2D latent spaces are compared across multiple linear and nonlinear dimensionality-reduction techniques (excluding autoencoders): \textbf{(a)} PCA, \textbf{(b)} t-SNE, \textbf{(c)} t-SNE following PCA, and \textbf{(d)} UMAP. The top row shows full latent spaces for a subset of the training data, while the bottom row highlights the same example sequence of 50 frames in each space. All but PCA show a very segmented and discontinuous path for the sequence across the latent space. Our autoencoder techniques (see Figure \ref{fig:ae_comparison}) are comparable to PCA in terms of continuity of the paths in latent space, but have a much higher capacity to learn complex, nonlinear relationships than PCA alone.}
\label{fig:pca}
\end{figure*}

While there are many other dimensional reduction techniques for data visualization, such as PCA, UMAP, and t-SNE \cite{pca,umap,tsne}, a significant advantage of autoencoders is that they learn a nonlinear mapping to the latent space that is by construction (approximately) invertible. Some differences between these other dimensionality-reducing techniques are illustrated in Figure \ref{fig:pca}. 

In principle, autoencoders can be used to synthesize new dance sequences by decoding any arbitrary trajectory through the latent space. We prioritize continuity and smoothness of paths in the latent space when possible, as this allows human-generated abstract trajectories (for example, traced on a phone or with a computer mouse) a greater likelihood of creating meaningful choreographies. These qualities of trajectories in the latent space are most prevalent in PCA and our autoencoder methods (see Figure \ref{fig:pca}). However, as PCA is a linear dimensionality-reduction method, it is far more limited in ability to conform to the full complexity of the realistic data manifold compared to autoencoder methods. 

The autoencoders' latent spaces do tend to produce mostly continuous trajectories for real sequences in the input data. This continuity can be greatly enhanced by subtracting out angular and positional orientation of the dancer, as shown in Figure \ref{fig:ae_comparison}. Removing these dimensions of variation further reduces the amount of information that must be stored by the autoencoder and allows it to create less convoluted mappings of similar poses regardless of the overall spatial orientation of the dancer.

However, absent a deliberate human-selected trajectory as an input, it is \emph{a priori} unclear how to select a meaningful trajectory, i.e., one that that corresponds to an aesthetically or artistically interesting synthetic performance.

\textls[19]{In order to address this limitation, and to give some insight into the space of ``interesting'' trajectories in the latent space, we take another approach in which a second autoencoder is trained to reconstruct fixed-length sequences of} 

\begin{figure*}[hb!]
\begin{minipage}{0.6\linewidth}
\includegraphics[width=\linewidth]{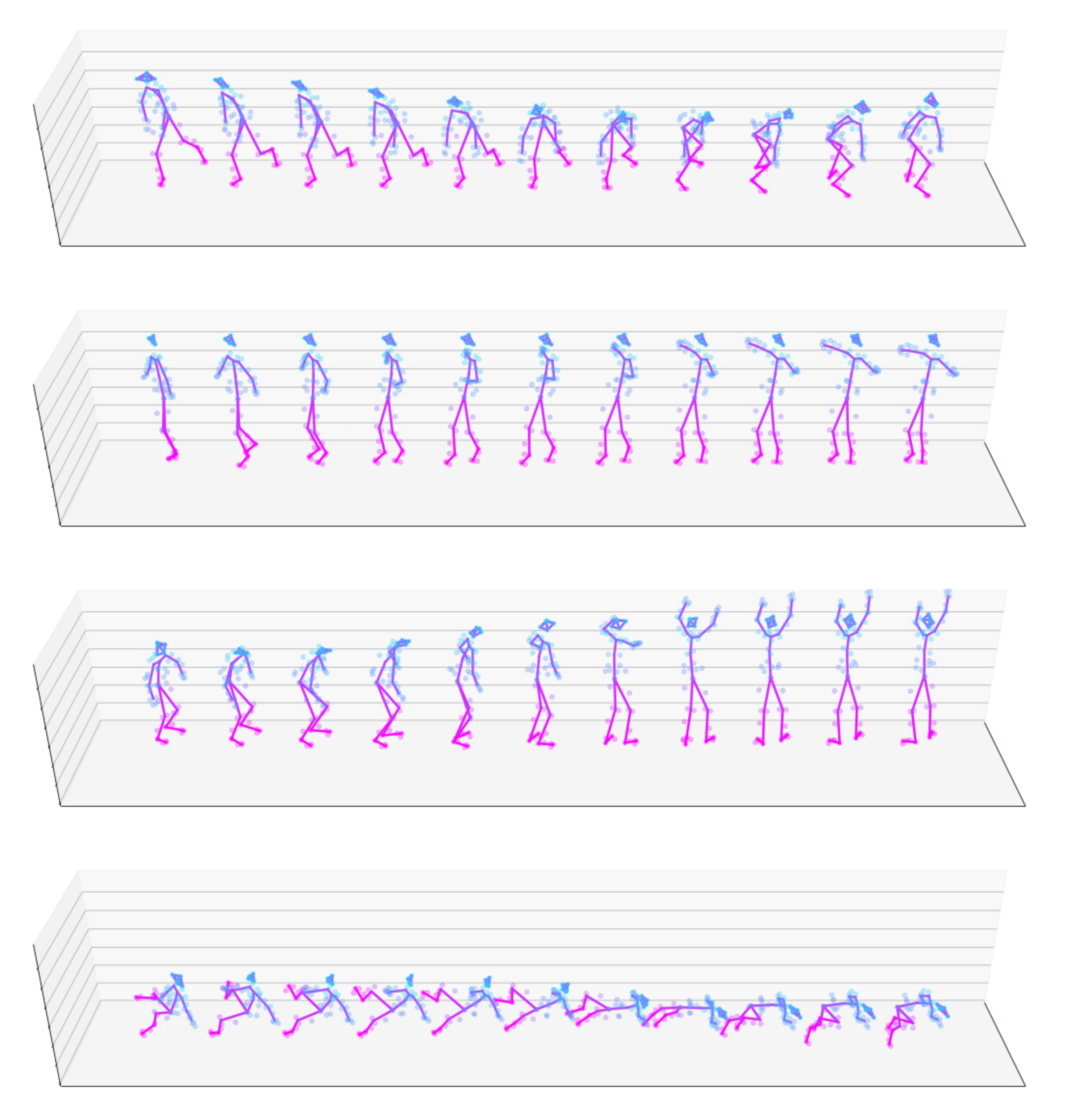}
\caption{Unconditionally-sampled sequences from the VAE.}
\label{fig:unconditional}
\end{minipage}
\begin{minipage}{0.4\linewidth}
\includegraphics[width=\linewidth]{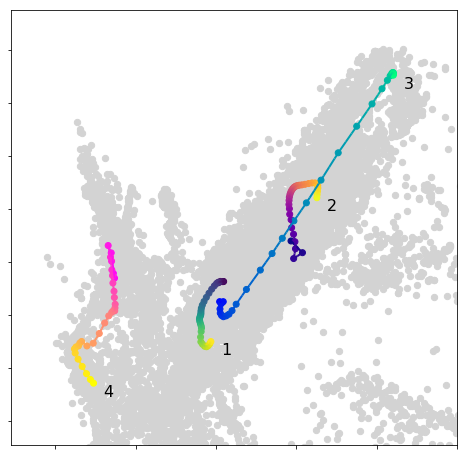}
\caption{Unconditionally-sampled sequences from the VAE projected into the latent space of the pose autoencoder (1 = top-most sequence; 4 = bottom-most sequence). Trajectories begin at darker colors and end at lighter colors.}
\label{fig:unconditional_paths}
\end{minipage}
\end{figure*}

\pagebreak

\noindent dance poses by mapping each sequence to a single point in a high-dimensional latent space. Moreover, we train this network as a Variational Autoencoder (VAE) \cite{vae} which attempts to learn a latent space whose distribution is compatible with a $(d\times m)$-dimensional Gaussian. Sampling from this latent space results in unconditionally-generated sequences that are realistic and inventive (see Figure \ref{fig:unconditional}). For each sampling, we look at a single point in the latent space corresponding to a fixed-length movement sequence. Within the scope of this paper, we do not attempt to impose any continuity requirements from one sampling to the next. Latent space points are chosen approximately isotropically. This creates a complementary creative tool to our previously-described traditional autoencoder for poses. We anticipate that choreographers and researchers could draw continuous paths through the latent space of poses to generate new movements as well as sample from the VAE latent space to generate new movement phrases and/or variations on existing phrases.

With both standard and variational autoencoders trained to replicate single poses and sequences of poses respectively, we introduce some techniques for taking a given input phrase of movement and generating infinitely many variations on that phrase. We define ``variation'' to mean that the overall spirit of the movement be preserved, but implemented with slightly different timing, intensity, or stylistic quality. 

After identifying a desired dance phrase from which to create variations, we identify the sequence of points in the latent space representing that sequence of poses. We first constructed trajectories close to the original sequence by adding small sinusoidal perturbations to the original sequence. This created sequences resembling the original phrase, but with an oscillatory frequency that was apparent in the output. This frequency could be tuned to the choreographer's desired setting, if the oscillatory effect is desired. However, we also sought out a method that constructed these paths in a less contrived manner.

For a VAE trained on sequences of poses, each point in the latent space represents an entire sequence of a fixed length $m$. We can construct variations on the input sequence by adding a small amount of random noise to the latent point and then applying the decoder to this new point in the latent space. This creates a new generated variation on the original sequence, with a level of ``originality'' that scales with the amount of noise added. Since the VAE's latent space has been constrained to resemble a Gaussian distribution, we can sample frequently from the latent space within several standard deviations of the origin without observing highly unphysical output sequences. Sampling within less than approximately $0.5\sigma$ tends to give very subtle variations, usually in timing or expressiveness in the phrase. Sampling within approximately $1\text{ to }2\sigma$ gives more inventive variations that deviate further from the original while often preserving some element of the original, e.g. a quick movement upwards of a limb or an overall rotational motion. Sampling within 3 to 4$\sigma$ and higher can produce myriad results ranging from no motion at all to extreme warping of the body to completely destroying the sense of a recognizeable human shape.

\begin{figure*}[!h]
\centering
\begin{minipage}{0.6\linewidth}
\centering
\includegraphics[width=0.9\linewidth]{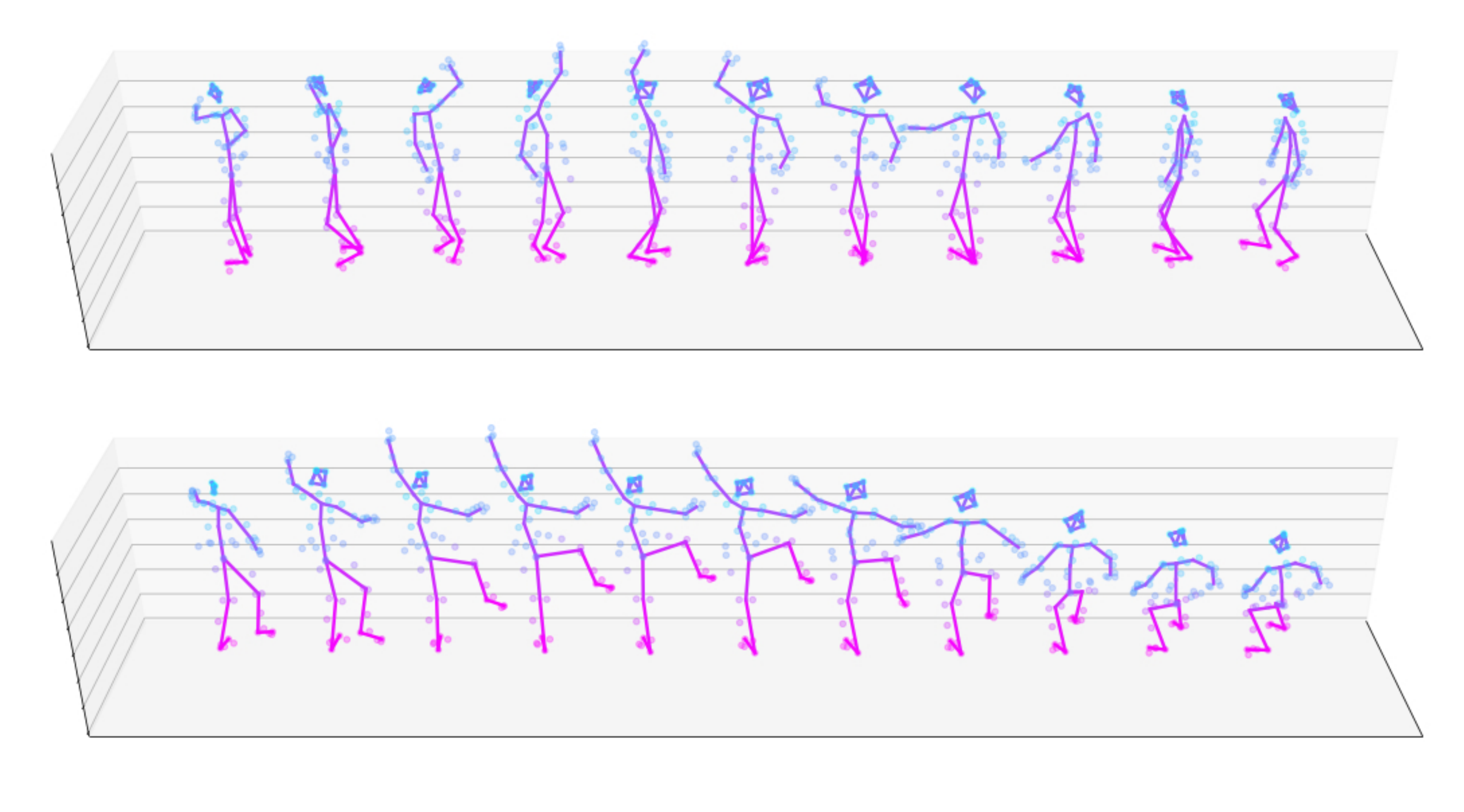}
\caption{A reference input sequence (above) and a generated variation sequence (below) with 0.5$\sigma$ noise added to the input's representation in latent space, both with lengths of 32 frames (time progressing from left to right). While the reference sequence includes a rotation, the generated variation removes the spin, while the movements of the left arm are synchronous in both cases.}
\label{fig:variation}
\end{minipage}
\hspace{1cm}
\begin{minipage}{0.3\linewidth}
\centering
\includegraphics[width=\linewidth]{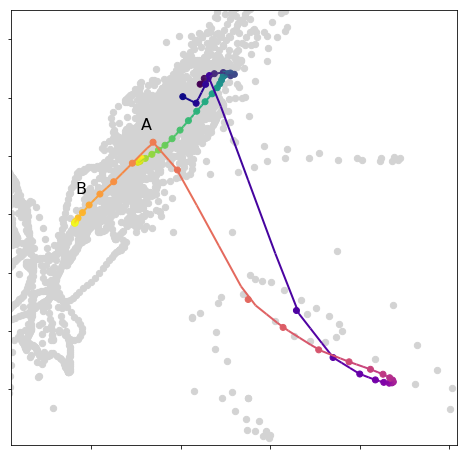}
\caption{Reference (A) and generated (B) variation sequences projected into the pose autoencoder space. Trajectory colors go from dark to light over time.}
\label{fig:variation_paths}
\end{minipage}
\begin{minipage}{0.6\linewidth}
\centering
\includegraphics[width=0.9\linewidth]{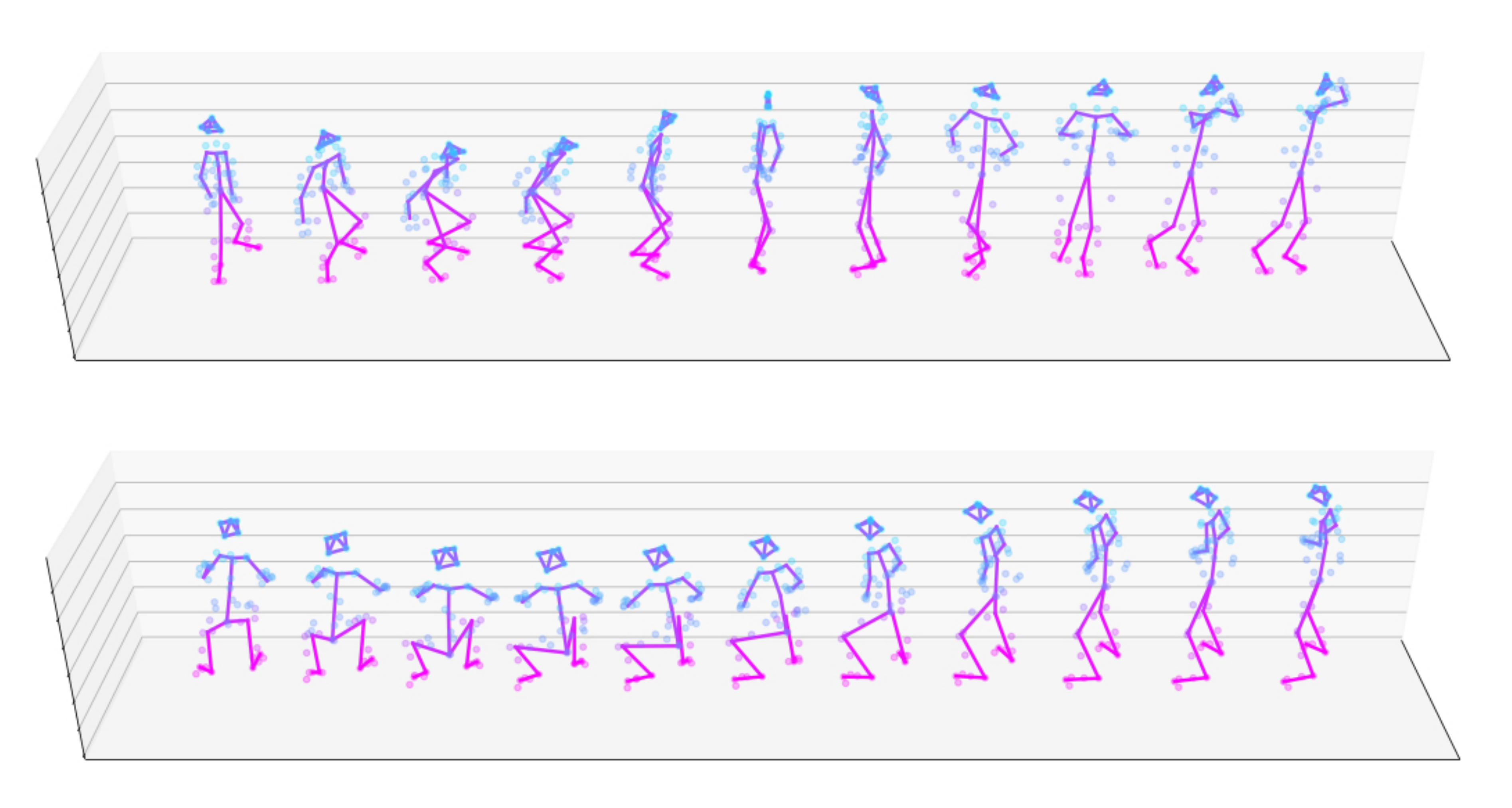}
\caption{A reference input sequence (above) and a generated variation sequence (below) with 0.5$\sigma$ noise added to the input's representation in latent space, both with lengths of 32 frames (time progressing from left to right). The generated variation preserves the rising motion but adds a rotation.}
\label{fig:variation2}
\end{minipage}
\hspace{1cm}
\begin{minipage}{0.3\linewidth}
\centering
\includegraphics[width=\linewidth]{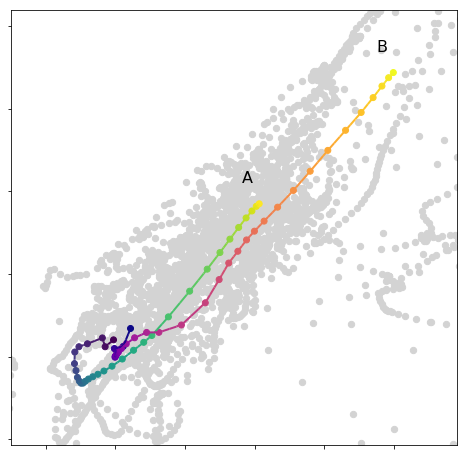}
\caption{Reference (A) and generated (B) variation sequences projected into the pose autoencoder space. Trajectory colors go from dark to light over time.}
\label{fig:variation2_paths}
\end{minipage}
\begin{minipage}{0.6\linewidth}
\centering
\includegraphics[width=0.9\linewidth]{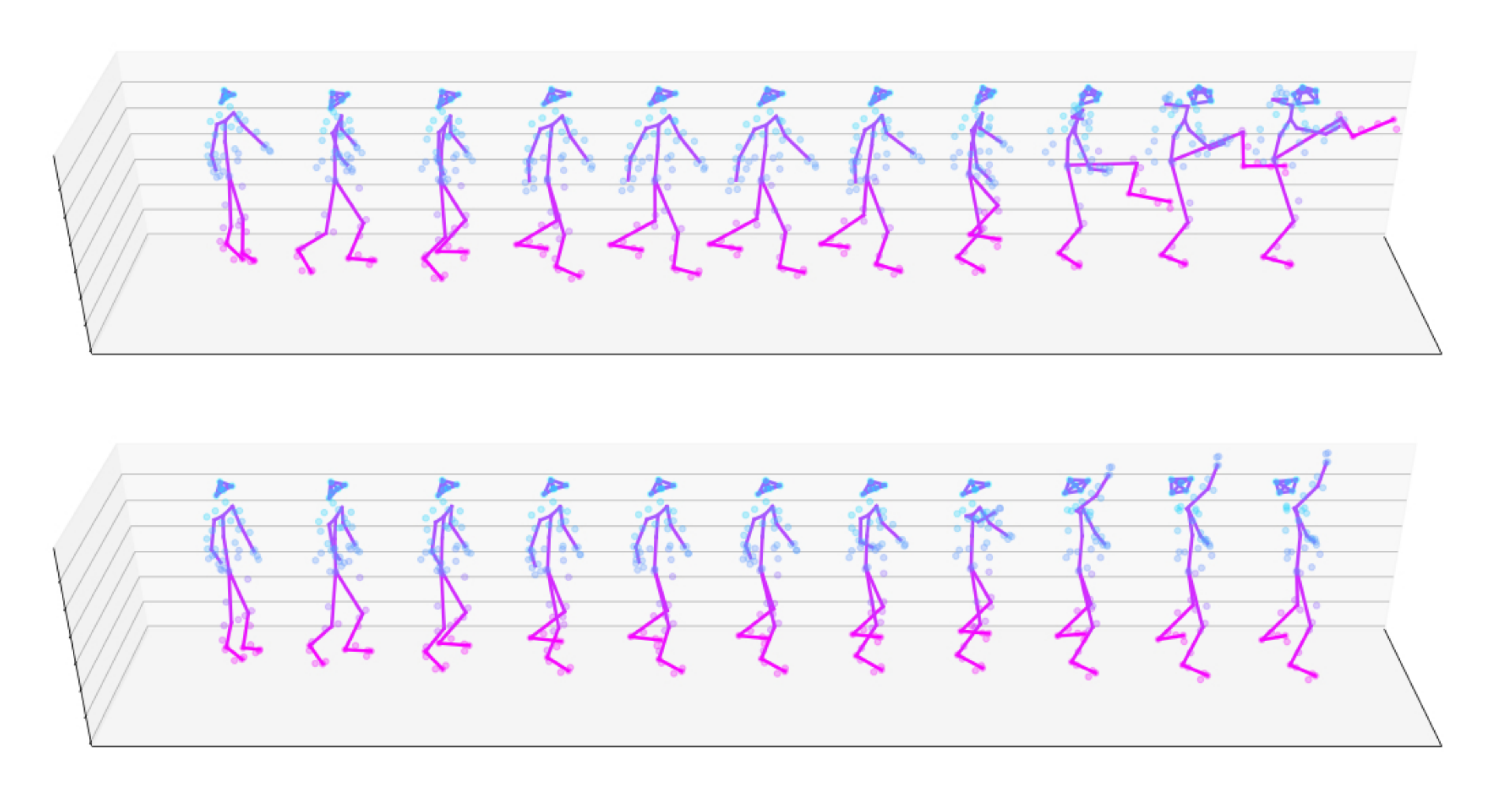}
\caption{A reference input sequence (above) and a generated variation sequence (below) with 0.5$\sigma$ noise added to the input's representation in latent space, both with lengths of 32 frames (time progressing from left to right). The reference sequence features a kick, while the variation instead translates this upward motion into the arms, rather than the feet.}
\label{fig:variation3}
\end{minipage}
\hspace{1cm}
\begin{minipage}{0.3\linewidth}
\centering
\includegraphics[width=\linewidth]{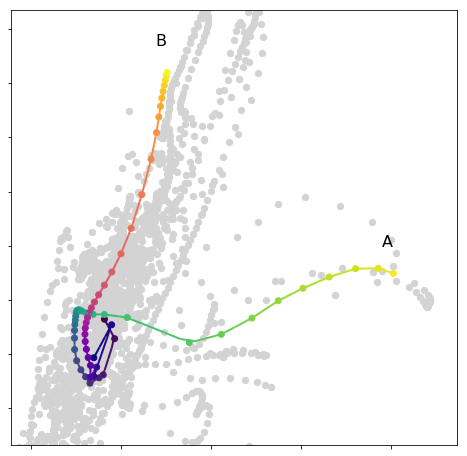}
\caption{Reference (A) and generated (B) variation sequences projected into the pose autoencoder space. Trajectory colors go from dark to light over time.}
\label{fig:variation3_paths}
\end{minipage}
\end{figure*}

The relationship between these two latent spaces -- that of the pose autoencoder and that of the sequence VAE -- may be exploited to gain insight into the variations themselves. Points in the VAE latent space directly map to trajectories in the pose autoencoder space. By introducing a slight amount of noise to the point in the VAE latent space corresponding to a desired input sequence, we may decode nearby points to construct trajectories in pose space that are highly related to the original input sequence. Examples of variations from reference sequences are shown in Figures \ref{fig:variation} - \ref{fig:variation3_paths}.

\section{Results and Discussion}

Both the RNN+MDN and autoencoded outputs created smooth and authentic-looking movements. Animations of input and output sequences for various combinations of our model parameters may be viewed here: \url{http://www.beyondimitation.com}. 

Training the RNN+MDN with a PCA dimensionality reduction tended to improve the quality of the generated outputs, at least in terms of the reconstruction of a realistic human body. We used PCA to transform the input dataset into a lower-dimensional format that explains 95\% of its variance. This transformation of the training data shortened the training time for each epoch by up to 15\%, though test accuracy was not significantly affected. The output resulted in a realistic human form earlier in the training than without the application of PCA. In the future, we may also investigate nonlinear forms of dimensionality reduction to further improve this technique.

The architectures used for the RNN+MDN models included 3 LSTM layers with sizes varying from 32 to 512 nodes. They took input sequences of length $m$ ranging from 10 to 128 and predicted the following $n$ frames ranging from 1 to 4 with a learning rate of 0.00001 and the Adam optimizer \cite{adam}.

The final architecture for the pose autoencoder comprises an encoder and a decoder each with two layers of 64 nodes with LeakyReLU activation functions with $\alpha=0.2$ and compiled with the Adam optimizer. The pose autoencoder takes inputs of shape $(53\times3)$ and maps them into a latent space of 32 dimensions. Training this over 80\% of our full dataset with a batch size of 128 and a learning rate of 0.0001 produced nearly-identical reconstructions of frames from the remaining 20\% of our data after about 50 epochs. We also trained a modification of this architecture with a data augmentation technique that added random offsets between $[0,1]$ to the $\hat{x}$ and $\hat{y}$ axes. This did not yield a significant advantage in terms of test accuracy, however, so we did not use it for our latent space explorations.

The final architecture for the sequence VAE also comprises an encoder and a decoder, each with 3 LSTM layers with 384 nodes and 1 dense layer with 256 nodes and a ReLU activation function, where 256 represents the dimensionality of the latent space. The model was compiled with the Adam optimizer. The VAE maps inputs of shape $(53\times3\times l)$, where $l$ is the fixed length of the movement sequence, to the $(256\times l)$-dimensional latent space and then back to their original dimensionality. We used input sequences of length $l=128$, which corresponds to about 4 seconds of continuous \noindent\textls[30]{movement. We augmented our data by rotating the frames} 

\pagebreak

\noindent in each batch by a randomly-chosen $\theta\in[0,2\pi]$. The VAE was trained with a learning rate of 0.001, a Kullback-Leibler weight = $0.0001$, and a Mean Squared Error (MSE) loss scaled by the approximate resolution of the motion capture data for about 1 day on a CuDNN-enabled GPU. 

Sampling from the latent space of standard and variational autoencoders for both poses and sequences provided a rich playground of generative movements. We are particularly interested in the dynamic range provided by these tools to create variations on input sequences: by increasing the magnitude of the perturbation of the latent sequence to be decoded, choreographers can decide how `creative' the outputs should look. By opting for either a standard or variational autoencoder, choreographers can sample from latent spaces with a bit more or a bit less similarity in the movements themselves to the training data. Adding sinusoidal perturbations as well as generating stylistically-related variations by exploiting the relationship between these two latent spaces proved effective and compelling methods for creating choreographic variations. The subtlety and smoothness with which we can vary input sequences using the VAE also underscores that the model is truly generating new outputs rather than memorizing the input data. 

These methods have already been effective at sparking choreographic innovation in the studio. They center the choreographer's embodied knowledge as something to be modeled and investigated -- not just as a compendium of possible bodily positions, but as a complex and high-dimensional landscape from which to sample movements both familiar and foreign. Movements throughout these abstract landscapes can be constructed in a variety of ways depending on the application. For example:
\begin{itemize}
    \item For a choreographer seeking primarily to document their practice, training these models allows them to save not only the physical motions captured in the motion capture data, but also their \emph{potential} for movement creation as approximated by a well-trained model. Different models may be saved from various periods of their practice and compared or re-explored indefinitely.
    \item For a choreographer looking to construct a new piece out of their own typical patterns of movement, sampling from within $1\sigma$ in the VAE latent space can generate multiple natural-looking phrases that can then be stitched together in the studio to create a cohesive piece. They could also prompt new sequences of arbitrary length following from existing choreography via the RNN+MDN model.
    \item For a choreographer who wants to understand and perhaps break out of their typical movement patterns, analyzing the latent space of the pose autoencoder can be instructive. Visualizing trajectories through the space can inform what areas lie unexplored. Drawing continuous paths through the latent space can then construct new phrases that might otherwise never emerge from an improvisation session. 
    \item A choreographer might also use these methods to support teaching movements to others. By comparing trajectories in the same latent space, students can track their mastery of a given movement sequence.
\end{itemize}

These creative tools allow a mode of documentation that opens up valuable reflection in the recursive process of movement creation. Since a significant portion of the choreographic process can be kinesthetically driven, it is useful to be able to externalize movement into the visual domain in order to reflect on the architecture and design of the choreography. This resource may double as a limitation if dance-makers rely solely on the visual aspect of choreography. Just as a mirror can serve as a double-edged sword in dance practice, these tools make explicit the possibilities of differentiation between the internal, kinesthetic dimension of movement research and the external, visual one. 

Generating novel movement allows us to see potential flow of choreographic patterns, which makes negotiating the aesthetic dimension richer if we take the time to evaluate why something looks subjectively unnatural. In this way, a dance-maker has a chance to articulate more clearly their own aesthetic preferences for choreographic intention and execution. 

As our title suggests, ``beyond imitation'' also points to the important distinction between creative expression and research-based inquiry. While dance-making certainly involves generation that is spontaneous and intuitive, choreographers may also take years honing and developing sequences that are deeply textured and multi-faceted. Disrupting any implicit hierarchies, these tools enable documentation of the systematic, recursive process of dance-making that is often so invisible and mysterious.

Future technical work to develop these methods will include the investigation of nonlinear, invertible data-reduction techniques as a form of pre-processing our inputs, other neural network-based models designed to work with timeseries data such as Temporal Convolutional Networks, and more sophisticated methods for sampling from latent spaces.

We can also increase the size of our training dataset by sourcing data not only from motion capture sessions, but also using OpenPose software to extract pose information from dance videos or even a laptop camera \cite{openpose,billtjonesai}. This could open up a provocative path in machine-augmented choreography: generating movements in the styles of any number of prominent choreographers.

Feedback from other choreographers who used our interactive models also indicated that it would be interesting to extend our current dataset with additonal data focused on the isolation of certain regions of the body and/or modalities of movement. Our next steps in extending this work will also include exploring latent spaces of multiple dancers. While only solo dances were captured for the studies in this paper, the Vicon system can readily accommodate multiple simultaneous dancers, which will allow us to explore the generation of duets and group choreographies.

\section*{Acknowledgements}
We would like to thank Raymond Pinto for many useful conversations and insights as well as providing additional movement data for future studies. This work has been supported by the generosity of the Yale Center for Collaborative Arts \& Media, the Yale Womens' Faculty Forum, and the Yale Digital Humanities Lab.

\vfill

\pagebreak

\section*{Appendix A: Mixture Density Networks}
The structure of a Mixture Density Network, as laid out in detail in \cite{bishop}, allows us to sample our target predictions from a linear combination of $m$ Gaussian distributions, each multiplied by an overall factor of $\alpha_i$, rather than from a single Gaussian. The probability density is therefore represented by 

\begin{equation*}
    p\left(\vec{t}\ |\ \vec{x}\right) = \sum_{i=1}^m \alpha_i(\vec{x}) \phi_i \left(\vec{t}\ |\ \vec{x}\right)
\end{equation*}

where $\vec{x}$ represents our input data, $\vec{t}$ reprents a given predicted output, $m$ represents the total number of Gaussian distributions in the mixture, and $c$ represents the total number of components to predict (here, $53\times 3$ for each timeslice). Each of the Gaussian distributions is modeled as: 

\begin{equation*}
    \phi_i\left(\vec{t}\ |\ \vec{x}\right) = \frac{1}{(2\pi)^{\frac{c}{2}}\sigma_i(\vec{x})^c} e^{-\frac{|\vec{t}-\vec{\mu}_i(\vec{x})|^2}{2\sigma_i(\vec{x})^2}}
\end{equation*}

Here, $\vec{\mu}_i(\vec{x})$ and $\sigma_i(\vec{x})$ represent the mean values and variances for each component of the generated output.

\bibliographystyle{iccc}
\bibliography{iccc}

\end{document}